%% file: main_arxiv.tex
\documentclass{article} 
\usepackage{arxiv}
\input{math_commands.tex}

\usepackage{hyperref}
\usepackage{url}
\usepackage[utf8]{inputenc} 
\usepackage[T1]{fontenc}    
\usepackage{booktabs}       
\usepackage{amsfonts}       
\usepackage{nicefrac}       
\usepackage{microtype}      
\usepackage{xcolor}         
\usepackage{makecell}
\usepackage{ifthen}
\usepackage{amsmath}
\usepackage{amsthm}
\usepackage{algorithm}
\usepackage{algpseudocode}
\usepackage{graphicx}
\usepackage{subfigure}
\usepackage{cleveref}
\usepackage{multicol}
\usepackage{natbib}

\usepackage{caption}

\newtheorem{lemma}{Lemma}

\newtheorem{proposition}{Proposition}
\newtheorem{corollary}{Corollary}

\newtheorem{remark}{Remark}

\newcommand{\blue}[1]{{\color{blue} #1}}

\title{\centering Joint Encoding of KV-Cache Blocks for Scalable LLM Serving}

\author{%
  Joseph Kampeas \& Emir Haleva \\
  Distributed and Parallel Software Lab\\
  Huawei Technologies \\
  \texttt{first.last@huawei.com} \\
  }

\begin{document}

\maketitle

\begin{abstract}
Modern large language models (LLMs) drive interactive AI systems but are bottlenecked by the memory-heavy growth of key-value (KV) caches, which limits real-time throughput under concurrent loads. Existing KV-cache compression methods rely on rigid heuristics, disrupt tensor layouts, or require specialized compute, hindering scalability and deployment. 

We propose joint encoding of KV-cache blocks, which fuses similar blocks across requests and input chunks into shared representations while preserving standard cache structure. This alleviates the KV-cache memory bottleneck, supporting high-concurrency serving without specialized hardware. Theoretically, we analyze the rate-distortion tradeoff of fused cache blocks under a Poisson process model. Empirically, our method achieves up to 4.38× KV-cache compression with negligible accuracy loss across diverse LLMs and benchmarks, outperforming recent structured and adaptive compression baselines. In real LLM serving, joint encoding improves the token throughput by $\sim$40\% on a single-machine vLLM benchmark, demonstrating substantial gains in inference throughput. Code is available at 
\href{https://github.com/sef1/kv_fast_fusion}{\nolinkurl{kv_joint_encoding}}. 
\end{abstract}

\input{sections/introduction_s}
\input{sections/related_s}
\input{sections/motivation}

\input{sections/method_s}

\input{sections/analysis}
\input{sections/results}

\section{Conclusion}\label{sec.conclusion}
In this paper, we presented \emph{Fast Fusion}, a novel context-sharing enhancement that can improve LLM serving efficiency by introducing BFF and CFF. These techniques enable fine-grained fusion of similar KV cache blocks across requests or chunks, achieving up to $\times$4.38 compression without compromising accuracy. By significantly reducing KV cache transfers, our method reduces significantly the average number of blocks, and thus, allows to scale the serving capacity effectively under heterogeneous workloads. Theoretical analysis based on Poisson point processes provides insight into the rate-distortion trade-offs, and extensive empirical evaluations across multiple benchmarks and model sizes validate the practical benefits. Looking forward, we plan to explore the impact on serving throughput, using adaptive threshold tuning, integration with quantization, and pruning methods.




\appendix
\input{sections/appendix}

\end{document}

%% file: math_commands.tex
\usepackage{bm,dsfont}

\def\1{\bm{1}}

\newcommand{\prnt}[1]{\left(#1\right)}

\newcommand{\brcs}[1]{\left\{#1\right\}}

\DeclareMathAlphabet{\mathsfit}{\encodingdefault}{\sfdefault}{m}{sl}
\SetMathAlphabet{\mathsfit}{bold}{\encodingdefault}{\sfdefault}{bx}{n}



%% file: sections/introduction_s.tex
\section{Introduction}

Large Language Model (LLM) services have recently gained tremendous popularity, but their serving pipeline faces distinct challenges across two stages. In the \emph{prefill stage}, the model processes the input sequence. This stage is compute-bound but executed only once per request. In contrast, the \emph{decode stage} generates the subsequent output tokens, is repeated for every token, and is primarily constrained by memory bandwidth. The bottleneck arises because each decoding step requires fetching large model parameters from memory to maintain fast computation \citep{xie2025reimagining}. This implies that batch size can be increased to serve more concurrent users without increasing generation latency, up to the point where computation time aligns with memory access time and becomes the new bottleneck.

Scaling LLM services for high concurrency rapidly encounters memory constraints. On-chip memory is consumed by three main components: the \emph{model weights}, the \emph{inference activations}, and the \emph{key–value (KV) cache}. The weights, though large, are static and do not grow with users concurrency. Activations are lightweight, scale linearly with the number of users, and vanish after each step. The KV-cache, however, is both persistent and expansive: it scales with the number of active sessions, and requires reserving memory proportional to the maximum sequence length per user. In practice, the KV-cache rapidly becomes the dominant memory consumer, and hence, it is the primary obstacle to scaling users concurrency.

To alleviate these memory constraints, recent serving frameworks borrow ideas from virtual memory and paging in operating systems. In particular, vLLM introduces \emph{Paged-Attention (PA)} \citet{kwon2023vllm}, which partitions the KV-cache of each request into fixed-size blocks, each holding a predefined number of key-value tokens. These blocks do not need to be contiguous in memory; instead, a \emph{block table} tracks their locations, allowing the attention mechanism to retrieve relevant entries efficiently \citep{kwon2023efficient}. This design removes the need of reserving per-user memory proportional to the maximum sequence length. Instead, memory consumption scales with the actual demand, which is typically far smaller on average.

Although Paged-Attention mitigates memory fragmentation, it does not address the fundamental asymmetry between the compute-bound prefill stage and the memory-bound decode stage. To address this, the vLLM scheduler prioritizes \emph{prefill requests}, reducing the Time-To-First-Token (TTFT) and allowing larger decode batches. Yet, this prioritization comes at a cost of delaying \emph{decode requests}, increasing the Time-Between-Tokens (TBT). In long-context scenarios, this elevated TBT can become severe, stalling generation for several seconds \citep{agrawal2023sarathi}. Recently, studies suggested to separate the two stages: requests are first processed on prefill servers, and their KV-cache is then migrated to decode servers. This separation enables independent optimization of the stages, reducing TTFT during prefill and TBT during decoding \citep{qin2024mooncake,zhong2024distserve,hu2024inference}.

In the prefill phase, strategies such as \emph{Tensor Parallelism} and \emph{chunked-prefilling} have been shown to reduce TTFT and increase compute utilization \citep{agrawal2023sarathi,agrawal2024mnemosyne}. In the decode phase, however, compute utilization remains low because each step requires fetching the model parameters to generate only a single token per request. Increasing the batch size can mitigate this inefficiency, but larger batches demand proportionally more KV-cache and are limited by the memory.

To reduce this overhead, prefix sharing has been proposed: grouping requests with identical prefixes so they can share the cached blocks and hence minimize memory \citep{zhu2024relayattention,juravsky2024hydragen}. Yet, exact prefix matches are rare in heterogeneous workloads, limiting its practical benefit. This limitation motivates the need for more general block-sharing mechanisms that extend the advantages of prefix sharing without relying on exact prefix matches.

In this paper, we address the KV-cache bottleneck in LLM serving by introducing a \emph{joint-encoding scheme} that compresses and reuses similar cache blocks across requests. We develop two complementary methods: \emph{Batch Fast-Fusion (BFF)}, which fuses blocks \emph{across different requests} prior to decoding, and \emph{Chunks Fast-Fusion (CFF)}, which fuses blocks \emph{across input chunks} during prefilling. Both schemes reduce KV-cache size, enable larger batch sizes, and lower network bandwidth demands. At the core of our approach is a \emph{tree-structured fusion strategy} that efficiently identifies encoding opportunities while preserving model accuracy. By applying joint encoding in both prefill and decode phases, our scheme delivers high throughput even under heterogeneous workloads with diverse input/output lengths and arrival patterns. These new fusion strategies substantially extend the generality and efficiency of cache sharing, delivering both memory and bandwidth savings in heterogeneous, real-world workloads. 

The remainder of this paper is organized as follows: \Cref{sec.related} discusses the most relevant related work. \Cref{sec.motivation} outlines the motivation behind block fusion in LLM serving systems. \Cref{sec.bff_cff} describes the proposed scheme in detail. \Cref{sec.analysis} analyzes the scheme in lens of the point processes, \Cref{sec.results} presents the experimental results and evaluation, and \Cref{sec.conclusion} concludes the paper.

%% file: sections/related_s.tex
\section{Related Work}\label{sec.related}

A major line of research has focused on \emph{reducing the memory footprint of the KV-cache}, primarily through compression. Key approaches include quantization, low-rank approximation, and selective eviction. \emph{Quantization-based methods}, such as \citet{hooper2024kvquant, liu2024kivi}, compress key and value embeddings to 2–3 bits with minimal accuracy loss, substantially increasing batch size and throughput over standard precision. Low-rank and latent representations, for example,  Multi-Head Latent Attention (MLA) \citet{liu2024deepseek} and ReCalKV \citet{yan2025recalkv}  stored KV matrices in compact subspaces that are later reconstructed as needed. This approach achieves a significant reduction in the KV-cache size while maintaining attention accuracy. Recently, \citet{meng2025transmla} suggested a method for transforming the attention of pretrained  models to MLA architectures.

Given the redundancy of KV states \emph{across adjacent layers}, other works interpolate vector directions or use SVD to compress cross-layer states (e.g., \citet{liu2024minicache, chang2025xkv}), often retaining only a subset of tokens for quality. Combining quantization with these techniques achieves KV-cache compression ratios up to 5×. Recent work leverages adaptive arithmetic encoding \citep{liu2024cachegen}, head-level token importance with residual merging \citep{liu2025zsmerge}, and sensitivity-driven dynamic sparsity \citep{zhang2024unifying,yang2025kvlink} to further boost cache efficiency.

A complementary research direction explores \emph{memory sharing across requests} by reusing computation for common input segments. Frameworks like HydraGen \citep{juravsky2024hydragen}, RelayAttention \citep{zhu2024relayattention}, and vLLM/SGLang \citep{wu2025know, zheng2023efficiently} exploit prefix sharing, where requests with identical prefixes share cached states, reducing both computation and memory requirements. However, such methods are limited by the rarity of exact prefix matches in practical workloads. Fragmentation and fine-grained reuse of overlapping segments are considered in \citep{zhang2025jenga, prabhu2025vattention, yang2025kvlink}.

Despite significant progress, most prior methods rely on rigid heuristics, disrupt native tensor layouts, or require exact prefix matches, limiting scalability and flexible integration into modern LLM serving pipelines. Our work directly addresses these limitations by introducing joint compression of KV-cache blocks, allowing finer-grained, layout-preserving, and general block sharing that extends the benefits of prefix reuse to arbitrarily similar segments. This flexibility is critical for memory- and bandwidth-efficient serving in real-world  settings.

%% file: sections/motivation.tex
\section{Motivation}\label{sec.motivation}
Improving the throughput of LLM serving is often achieved by increasing batch size. Since the decoding phase is primarily memory-bound, dominated by repeated fetching of large weight matrices, larger batch sizes do not substantially increase per-request decoding latency. However, serving scalability remains constrained by the memory footprint of the KV-cache, which must be reserved per user session. 
Apart from the common compression methods like quantization and sparsification (by eviction), prefix-sharing techniques 
like \citet{juravsky2024hydragen} and \citet{zheng2024batchllm}
demonstrate that overlapping prompt segments across requests enable KV-cache sharing. In addition, reuse of computed blocks as well as improved computation can improve throughput by up to 36\% \citep{jeong2024automatic}. Nonetheless, these approaches rely on \emph{exact} prefix matches, which severely limits applicability in real-world settings where variations in phrasing or task-specific inputs are common \citep{wu2025know, zheng2024batchllm}. For example, two translation requests with slightly different introductory phrases (“Help me translate” vs. “Translate this”) would fail to share any KV blocks under strict prefix matching \citep{wu2025know}.

\begin{figure}[t]
    \centering
    \subfigure[]{
        \includegraphics[width=0.45\textwidth]{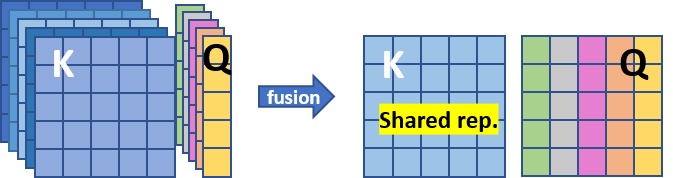}
        \label{fig:shared rep}
    }   
    \subfigure[]{
        \includegraphics[width=0.25\textwidth]{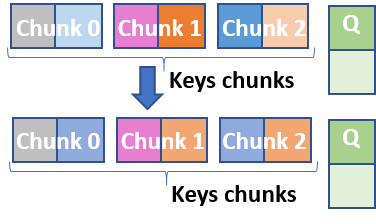}
        \label{fig:shared chunks}
   }
   
    \caption{Attention computation (a)  batch of requests with similar keys blocks. After the fusion, the KV=cache footprint is smaller and batch decoding is optimized by performing matrix-matrix multiplication. (b) After fusing chunks into a unified representation,  the blocks computation of chunks 0 and 1 can be reused in chunk 2 (same color). }
    \label{fig:sharing benefits}
\end{figure}

To overcome the strict requirement of an exact match, we propose a KV-cache blocks joint-encoding scheme that fuses blocks based on similarity  threshold. This approach extends the benefits of prefix sharing to non-identical contexts. \Cref{fig:sharing benefits} highlights the memory and computational benefits of jointly encoding blocks in Softmax Attention (SA).  Specifically, \Cref{fig:shared rep} illustrates the benefits in the decoding phase, where keys with similar representations (indicated by color) can be combined into a unified representation. Fusing these keys (and values) into a unified representation not only allows increasing the batch size but also facilitates optimized matrix-matrix multiplication, instead of multiple matrix-vector multiplications.
In the prefill phase, \Cref{fig:shared chunks} illustrates how chunking input requests allows joint encoding of blocks across chunks, which reduces KV blocks memory, and thus, facilitates handling longer inputs and further reusing computation in SA. Consequently, any SA computation involving a fused block can be reused in subsequent chunks that include instances of this block, maximizing efficiency and throughput. 

In both cases, after encoding, some blocks share the same representation and the block table points to the shared representation. Thus, any SA computation that involves a fused block can be reused in later chunks that comprise an instance of this block. The final SA result is derived by merging the SA of the fused components with that of the unique components (and rescaling). The recent version of vLLM can take advantage of the computation of SA with shared blocks. 

%% file: sections/method_s.tex
\section{KV Blocks Joint-Encoding}\label{sec.bff_cff}
In this section, we introduce our KV blocks joint-encoding scheme for the chunk-prefilling setting, to further facilitate the joint encoding of the chunks while they are processed in the LLM \citep{qin2024mooncake, agrawal2023sarathi}. The main objective of our scheme is to maximize the batch size by jointly encoding KV blocks. The impact is threefold.  First, in the decode phase, benefit from a higher arithmetic efficiency due to a larger batch size, and further, minimize KV-cache prefetching, mitigating the memory bandwidth limit. Second, in the prefill phase, it also enables the reuse of fused blocks computations, alleviating the compute limitation (see \Cref{fig:sharing benefits}). Third, reduce the usage of network bandwidth, which often becomes a bottleneck, especially when considering a high degree of parallelism~\citep{agrawal2024mnemosyne}. Certainly, it is crucial to achieve this goal while maintaining the model's accuracy.

To fulfill this goal, we propose Fast-Fusion (FF) method that jointly encodes similar blocks into a unified representation if the cosine similarity between contexts of different blocks is above predefined threshold. The similarity threshold is substantial for both the resulting \emph{compression ratio} and \emph{model's accuracy}. To avoid excessive blocks comparisons overhead, we design an efficient \emph{tree-like fusion strategy} that scales as $O(B \log B)$ and $O(C \log C)$, where $B$ is the batch size and $C$ is the number of chunks, respectively. This strategy allows the parallelization of the fusion process at each level of the tree. Accordingly, the fusion can be done over blocks of different requests (i.e., BFF) or blocks of different chunks (i.e., CFF). 

In practice, the KV-cache layout in vLLM is 
\((B,p,t, h, d)\), where $p, t, h$, and $d$ are the number of blocks, tokens per block, number of heads, and the head embedding size, respectively. Before encoding, the KV-cache is unfolded into a convenient layout of blocks per request or chunk. Specifically, let $r= t\cdot h \cdot d$, and note that \( C = \lfloor p \cdot t / (\text{size of chunk})\rfloor\), we use a layout of \((B,p, r)\) and \((C, p/C, r)\) for the BFF and the CFF, respectively. The norm of each $r$ is stored to allow a proper rescaling of the fused blocks.   The algorithm is applied iteratively for every $N$ chunks or requests in each layer. 

\begin{figure}
\begin{minipage}{.56\textwidth}
\begin{algorithm}[H]
\small
\caption{Fast Fusion}\label{alg: fast fusion}
\begin{algorithmic}
\Require {KV-cache (k,v), threshold ($thr$)}
\If{len(k) == 1}
    \State   \textbf{return} normalize(k), normalize(v) 
\EndIf    
    \State \blue{(*tree left*)}
    \State $\text{k}_l, \text{v}_l \gets \mathrm{FastFusion(k[:len(k)//2], v[:len(v)//2]}, thr)$
    \State \blue{(*tree right*)}
    \State $\text{k}_r, \text{v}_r \gets \mathrm{FastFusion(k[len(k)//2:], v[len(v)//2:]}, thr)$  
    \State $\mathrm{sim} \gets (\text{k}_l \text{k}_r^{\top})$ \blue{(*similarity per block of $\text{k}_l$*)}
    \For{ each $row$ in sim}
    \State $idx_\text{fused} \gets \text{indices where } \mathrm{sim}(row) > thr $
    \If{$\vert idx_\text{fused} \vert > 0$}
        \State \blue{(*block fusion*)}
        \State $\text{k}_l(row) \gets$ normalize($\text{k}_l(row)$ + $\sum_{i \in idx_\text{fused}}\text{k}_r(i)$)  
        \State $\text{v}_l(row) \gets$ normalize($\text{v}_l(row)$ + $\sum_{i \in idx_\text{fused}}\text{v}_r(i)$)  
        \State evict fused blocks at $\text{k}_r(idx_{\text{fused}})$ and $\text{v}_r(idx_{\text{fused}})$
        \State \blue{(*for matrix-matrix efficiency*)}
    \State mark shared blocks at $\text{k}_l(row)$, and $\text{v}_l(row)$ 
    \EndIf
    \EndFor
    \State update block-table $bt$
    \State \textbf{return} cat([$\text{k}_l$ | $\text{k}_r$ non-shared blocks]),
    \State $\qquad \qquad$ cat([$\text{v}_l$ |  $\text{v}_r$ non-shared blocks])
\end{algorithmic}
\normalsize
\end{algorithm}
\end{minipage}
\begin{minipage}{.43\textwidth}
        \centering
        \includegraphics[width=\textwidth]{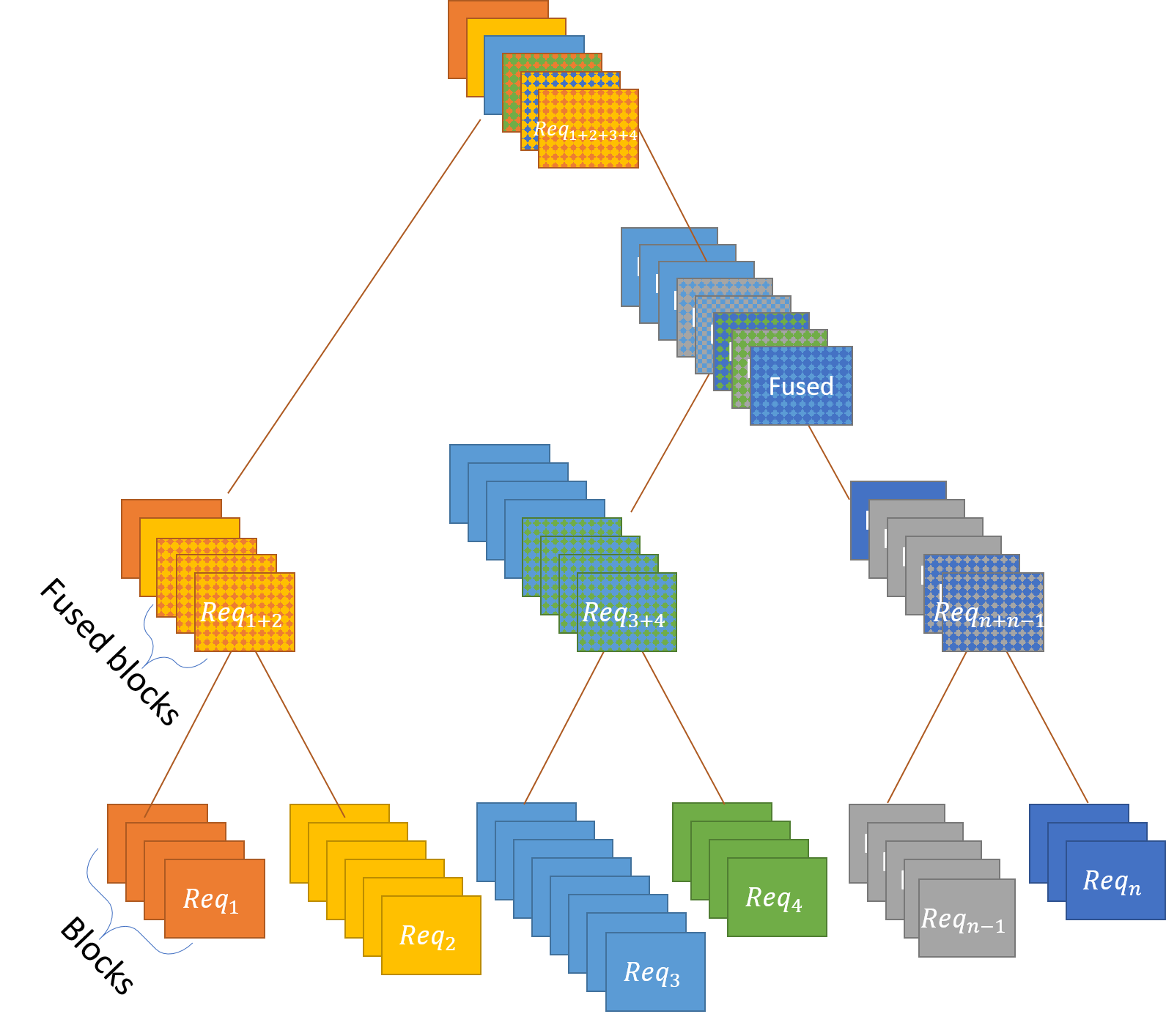}
        \captionof{figure}{BFF example for 6 request, where each request has a different number of blocks (and a different color). The Fast Fusion is recursively called, fusing every pair of requests (fused blocks are depicted by a mixed color). Then, fusing pairs of pairs, and so on, until all the  requests' blocks are jointly described in the KV-cache. Clearly, some blocks are not fused (remain their original color), and some are fused more than once while traversing the tree.}       
        \label{fig:bff example}
    
\end{minipage}
\end{figure}

A pseudo-code of the FF algorithm is given in \Cref{alg: fast fusion}. Roughly speaking, after  unfolding the KV-cache into a convenient layout and storing the norms, we recursively call the FF method in \Cref{alg: fast fusion}, which fuses blocks of different requests or chunks if their similarity level is above threshold. Intuitively, blocks represented by $r$-dimensional vectors can also be expressed in terms of their norm and corresponding $r$-dimensional unit (direction) vectors. In this view, fusion can be understood as aligning multiple unit vectors into a single unified direction, while \emph{preserving the distinctiveness of the original blocks through their norms}. This allows representing multiple blocks using a single unit vector and a norm (scalar) per block.  To further enhance the compute, the fused blocks are taken into account in the attention computation, 
allowing low-level kernels to leverage jointly encoded blocks in the decoding phase, and reuse computations in the prefill phase. Only one copy of the fused blocks is needed, and redundant copies can be evicted to reduce memory usage. Further,  different number of blocks may be encoded in each layer, which requires running each layer with its own block table, as implemented in vLLM. Note that our scheme attains a compression that is at least the compression attained by shared prefix methods, since we further compress inputs that do not share prefixes. 



Consequently, the joint encoding allows us to reuse computations to address the limitations of compute, memory bandwidth, and network bandwidth. The tree structure is beneficial for detecting fusion opportunities at a reasonable cost in large-scale prefill and decoding serving systems.  
\begin{remark}
    The order of requests impact the resulting compression. Specifically, it is beneficial to place shorter requests (in terms of number of blocks) on the left tree to gain higher diversity per block. At the best case,  all the blocks in the right tree will be unified into the left tree blocks representation, yielding the maximal compression.     
    However, the responsibility of ordering the requests belongs to the scheduler, which is out of the scope of this paper.  
\end{remark}
\begin{remark}    
    To avoid manual tuning, the threshold can be self-adjusted  using a simple feedback. For example, percentile-based or target-compression strategies monitor fusion statistics during serving: if too many blocks are fused (or compression exceeds a desired CR), the threshold is increased, and vice versa. Alternatively, a small calibration set can be used to refine the threshold based on acceptable distortion levels, providing a lightweight and fully automated adaptation mechanism.    
\end{remark}
\begin{remark}    
    Prior work (e.g., \citet{wu2025know}) demonstrated that prefix-sharing  may leak information through timing side-channels. In particular, by examining the TTFT, an attacker can determine whether a particular prefix has already been cached by the vLLM. Interestingly, joint encoding mitigates such risks: because \emph{multiple distinct KV blocks} are mapped to a shared latent direction, any reduction in compute time is no longer attributable to a single prefix or token, but to a many-to-one cluster. This substantially weakens the leakage signal compared to prefix matching.     
\end{remark}

%% file: sections/analysis.tex
\section{Analysis}\label{sec.analysis}
The similarity threshold is substantial to the algorithm performance as it dictates the resulting accuracy (or distortion) on the one hand, and the compression (or rate) on the other. In terms of rate-distortion, high threshold ensures low KV cache distortion, yet may result in low exceedance rate, and consequently, moderate  compression. Similarly, low threshold value allows unifying many blocks into a single representation (high compression), yet may degrade the performance due to a highly distorted KV cache.  Hence, it is crucial to thoroughly study the impact of the threshold.

\begin{figure*}[t]
\centering
 \subfigure[]{
 \centering
   \includegraphics[width=0.48\textwidth]{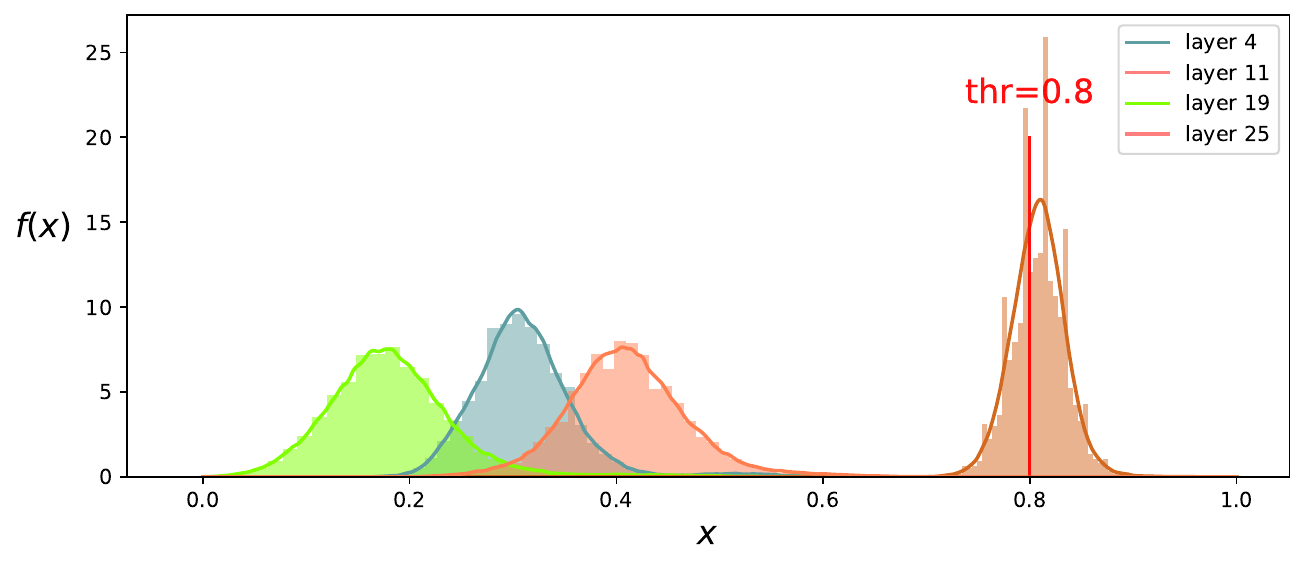}
     \label{fig:blocks similirty density}
 }
 \hfill
 \subfigure[]{
 \centering
     \includegraphics[width=0.475\textwidth]{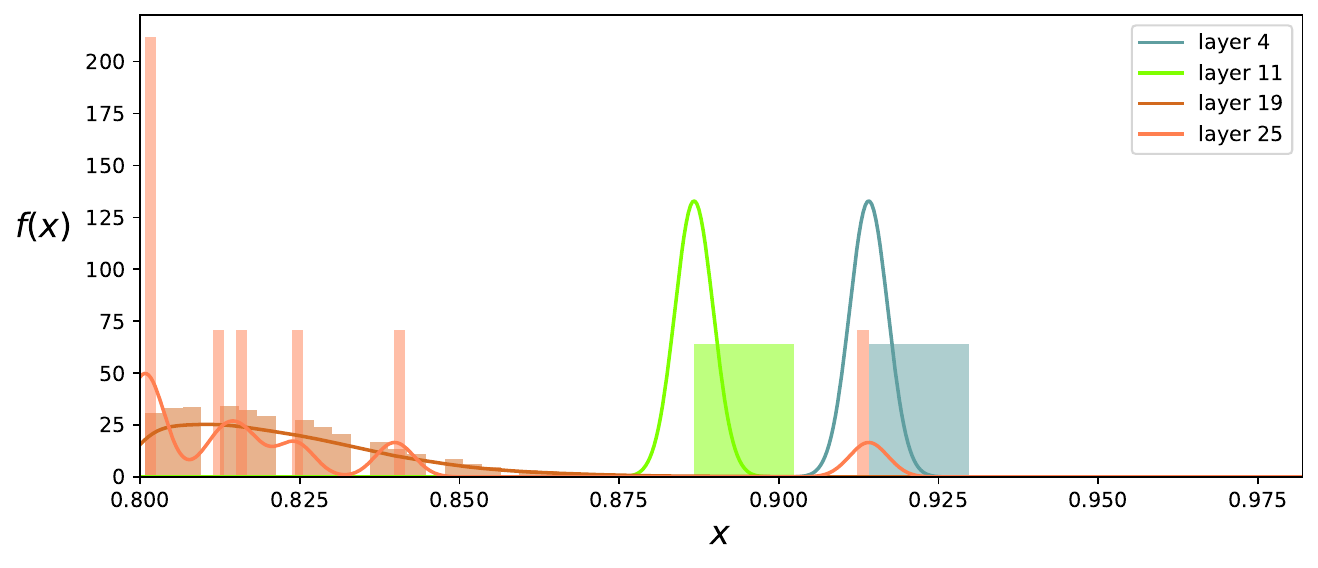}
      \label{fig:above-threshold similirty density}
 }
 \caption{Empirical similarity and analysis for \citet{DeepSeekR1DistillQwen7B} on \citet{nVidiaHelpSteer} dataset for several layers. Bars represents the empirical similarity and  solid line the KDE approximation. (a) Blocks similarity density.  (b) Above-threshold density for threshold=0.8.  }
 \label{fig:similarity analysis}
\end{figure*}

To analyze the resulting rate, recall that \Cref{alg: fast fusion} recursively fuses pairs of requests (see \Cref{fig:bff example} for illustration). Assume a pair of requests with $m_1$ and $m_2$ blocks, respectively, for which there are $n = m_1 \cdot m_2$ similarity samples \(\brcs{x_i}_{i=1}^n\). Then, using Kernel Density Estimation (KDE) with a Gaussian kernel \citep[Ch. 2]{wand1994kernel}, the similarity density in each layer is approximately 
\(
  f_h(x) = \frac{1}{nh} \sum_{i=1}^n\phi\left(\frac{x-x_i}{h}\right),  
  \label{eq: KDE density}
\)
where $h$ is the kernel bandwidth (standard deviation). Therefore, the similarity distribution is approximately
\begin{equation}
  F_h(x) = \frac{1}{n} \sum_{i=1}^n\Phi\left(\frac{x-x_i}{h}\right),
  \label{eq: KDE distribution}
\end{equation}
where $\phi$ and $\Phi$ are the Gaussian density and distribution functions, respectively. In other words, each sample $x_i$ contributes a Gaussian kernel function with mean value $x_i$ to the overall estimated probability density function, thus contributing a different probability of exceeding the threshold. 
\begin{proposition}\label{prop: poisson rate}
    For a sufficiently high similarity threshold $u$, the number of above-threshold similarities is asymptotically a Poisson variable with rate
    \begin{equation}\label{eq. lambda}
      \Lambda(u) = \frac{1}{n}\sum_{i=1}^n \exp\left\{-\frac{u - (h b_n + x_i)}{h a_n}  \right\},  
    \end{equation}    
where $a_n = (2\log n)^{-1/2}$ and $b_n = (2\log n)^{1/2} - \frac{1}{2}(2 \log n)^{-1/2}(\log \log n + \log(4\pi))$ are normalization constants.
\end{proposition}
\begin{proof}
Our goal is to analyze the asymptotic threshold exceedance rate of \cref{eq: KDE distribution}. Since the indices of high threshold exceedance are random and are relatively rare, the number of above-threshold observations can be modeled as a Poisson random variable when the threshold is sufficiently high \citep[Ch. 5]{leadbetter2012extremes}. In particular, let $u$ be a threshold such that the kernel of each sample $i$ satisfies \(\Pr(\mathbf{x}_i>u) = \prnt{1-\Phi\prnt{\frac{u - x_i}{h}}} =  \Theta(1/n)\). Then,  according to the \emph{uniformly asymptotically negligible} condition,  the number of threshold arrivals is approximately a Poisson variable with rate $\Lambda(u)$ when the samples are independent \citep[Ch. 8.3]{falk2010laws}. Similar treatment for dependent samples is given in \citep[Ch. 5]{coles2001introduction}.

For the Gaussian case, the threshold exceedance rate is \(\Lambda(u)\) given in \cref{eq. lambda},
where $a_n = (2\log n)^{-1/2}$ and $b_n = (2\log n)^{1/2} - \frac{1}{2}(2 \log n)^{-1/2}(\log \log n + \log(4\pi))$ are normalizing constants \citep[Theorem 5]{kampeas2014capacity}. 
\end{proof}
\Cref{fig:blocks similirty density} depicts the empirical similarity and the above-threshold distribution together with the KDE approximation for \citet{DeepSeekR1DistillQwen7B} on \citet{nVidiaHelpSteer} dataset in several layers. Evidently, for this model and dataset, the similarity appears Gaussian, yet with different mean and variance in each layer. 
Interestingly, letting the similarity in layer $\ell$ be approximately Gaussian with mean \(\mu_\ell = \frac{1}{n}\sum_{i=1}^n x_i\) and variance \(\sigma^2_\ell = \frac{1}{n}\sum_{i=1}^n x_i^2 - \mu_\ell^2\), by the Poisson point process,  the threshold exceedance rate is approximately \(\Lambda_\ell(u) = n \cdot \left(1-\Phi\left(\frac{u-\mu_\ell}{\sigma_\ell}\right)\right )\). Moreover, even though each layer follows a different distribution and, therefore, has a different threshold exceedance rate, by \Cref{prop: poisson rate},  the overall exceedance rate for all model layers can be evaluated. 

The Poisson formulation facilitates analyzing the probability of observing non-compressible layer, and the expected compression ratio over the entire model. From the properties of the Poisson, we have the following corollary. 
\begin{corollary}
    After the fusion, the compression ratio over $L$ layers is 
    \[\text{compression ratio} = L(m_1 + m_2)/\prnt{(L(m_1 + m_2) -  \sum_{\ell = 1}^ L \Lambda_\ell(u)}.\]
    The probability of no fusion in layer $\ell$ is
    \[\Pr(\text{no fusion in layer $\ell$}) = \exp(-\Lambda_\ell(u))\]
\end{corollary}
%
%

To analyze the resulting distortion, let us examine the attention distribution drift that stems from fusing blocks.
Let $\mathbf{z}_{t} = \mathbf{q} \cdot \mathbf{k}_{t} / \sqrt{d}, \forall t \in {1,\dots,p}$ be the logits vector of a block with $p$ tokens, and let let $\mathbf{s} = \exp(\mathbf{z})/\sum_j\exp({z}_j)$ be the softmax distribution before fusion. Similarly, let $\mathbf{z}^\prime = \mathbf{z} + \mathbf{\delta}$ and $\mathbf{s}^\prime = \exp(\mathbf{z}^\prime)/\sum_j\exp({z}^\prime_j)$ be the perturbed (fused) logits and softmax distribution. Then, the attention distribution drift satisfies the following bound. 
\begin{lemma}\label{lem:Attention (softmax) error bound}
After fusion, the attention distribution drift is bounded by 
\[
\Vert \mathbf{s}^\prime - \mathbf{s} \Vert_1 \leq 2\epsilon  + O(\epsilon^2) 
\]
where $\epsilon = \max_i \frac{\Vert \mathbf{k}_i \Vert \Vert \mathbf{q}\Vert}{\sqrt{d}} \sqrt{2(1-u)}$.
\end{lemma}
Proof is deferred to \Cref{proof:lem attn error}.
The analysis provides a theoretical foundation for understanding the trade-offs inherent in block fusion for LLM serving. The Poisson point process modeling indeed shows that the similarity threshold directly governs the balance between compression and distortion in the KV cache. Our per-layer analysis predicts the expected compression for a given threshold, allowing skipping non-compressible layers in probability. 

%% file: sections/results.tex
\section{Results}\label{sec.results}
In this section, we present the results of our fusion scheme for CFF and BFF. Measuring the accuracy, compression, and latency together requires a dedicated kernel, which we keep in mind for future work.  Sections \ref{sec.block diversity}-\ref{sec. cff prefill}  assess compression and accuracy, by fuse and replicate the rescaled cache across all fused blocks (without freeing blocks). \Cref{sec. e2e performance} examines the end-to-end latency by fusing and freeing blocks (without rescaling). 

We evaluate the performance of our approach using a commodity GPU on various benchmarks and models and compare it to the baseline performance. The results demonstrate the effectiveness of our scheduling algorithm in terms of compute efficiency, memory reduction, and network bandwidth consumption. 
\subsection{Block Diversity and Rate-Distortion}\label{sec.block diversity}
\begin{figure*}[t]
\centering
 \subfigure[]{
 \centering
   \includegraphics[width=0.31\textwidth]{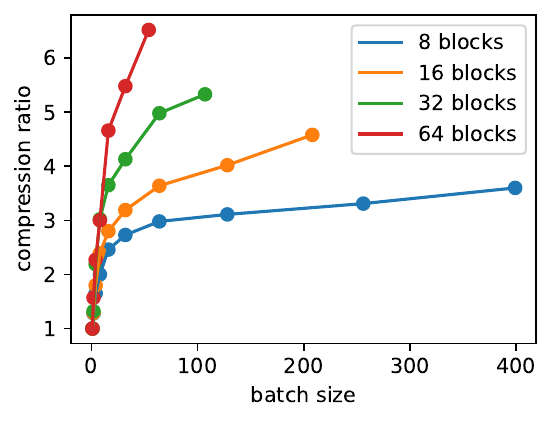}
     \label{fig:bff_random}
 }
 \hfill
 \subfigure[]{
 \centering
     \includegraphics[width=0.31\textwidth]{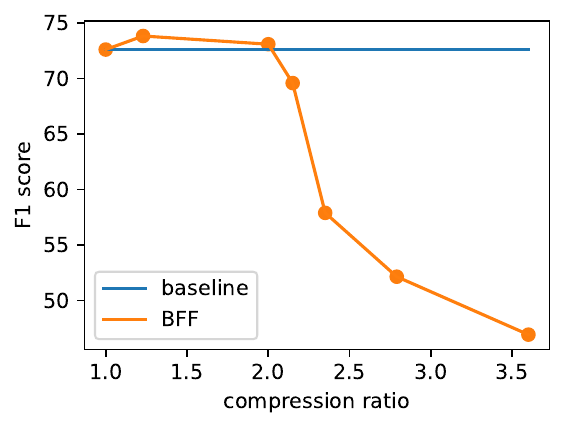}
      \label{fig:bff_rate_dist_therapy}
 }
  \hfill
 \subfigure[]{
 \centering
     \includegraphics[width=0.31\textwidth]{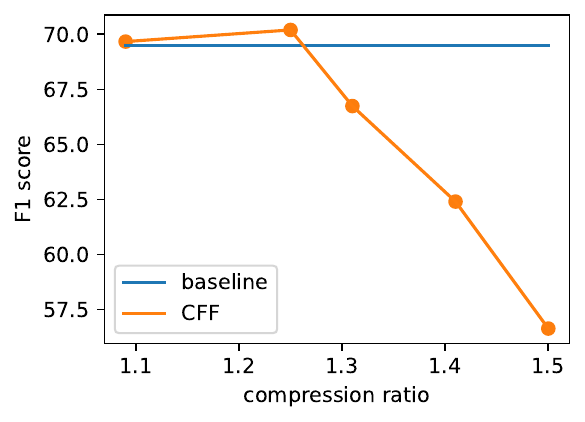}
      \label{fig:cff_rate_dist_therapy}
 }
 \caption{CR and F1 score of BFF and CFF for Llama-2 7B. (a) CR vs. batch size for diverse number of blocks on vLLM random-data benchmark. (b) BFF F1 score vs. CR for batch size 256  on conversational dataset. (c) CFF F1 score vs. CR, for 4 chunks on conversational dataset.}
 \label{fig:rate_dist_and_diversity}
\end{figure*}

Both the number of requests (or chunks) and the number of blocks in each request (or chunk) influence the resulting Compression Ratio (CR). The larger the batch, the more requests to fuse, and the more blocks per request, the greater diversity of blocks for fusion, which yields a better CR. To assess the significance of each, we examine the BFF CR on the random data benchmark in \citet{vLLMrandom}, using a fixed similarity threshold for a various number of requests and blocks per request using Llama2-7B with 16 tokens per block. Interestingly, \Cref{fig:bff_random} reveals that the diversity of blocks per request is more significant than the diversity of requests, since the CR grows faster when more blocks per request are used. 

\Cref{fig:bff_rate_dist_therapy} and \ref{fig:cff_rate_dist_therapy} depict the  the rate distortion trade-off for BFF and CFF, respectively. The rate distortion curve is given for Llama2-7B on \cite{Mr-Bhaskar_Synthetic} conversational dataset, where the number of requests (chunks) is fixed, and the similarity threshold varies. The baseline result is given for reference.   Specifically, in \Cref{fig:bff_rate_dist_therapy} the BFF achieves a CR of $\sim 2.15 \times$, without degrading accuracy  for batch size 256 with a block size of 16 tokens.  In \Cref{fig:cff_rate_dist_therapy}, the CFF achieves a CR of $\sim 1.25 \times$ without losing accuracy for 4 chunks with a block size of 16 tokens.  

Remarkably, this rate distortion formulation indicates that a higher CR can be achieved without sacrificing accuracy when fixing the similarity threshold value. In particular, since only the threshold determines the resulting accuracy (distortion), once the threshold is fixed, increasing the batch size or the number of blocks per request yields a better CR due to diversity in requests and blocks.  

\subsection{Batch Fast-Fusion During Decoding}\label{sec. bff decoding}
The impact of diversity on the CR and the resulting F1 score is of great practical interest, as it indicates the gain of a larger batch using our enhancement. Specifically, in this section, we investigate the CR and the F1 score when increasing the number of requests in the BFF scheme, where the similarity threshold is set to a fixed value, using the  \cite{nVidiaOpenMathInstruct} dataset. 

\begin{figure}[t]
    \centering
    \subfigure[]{
        \includegraphics[width=0.45\textwidth]{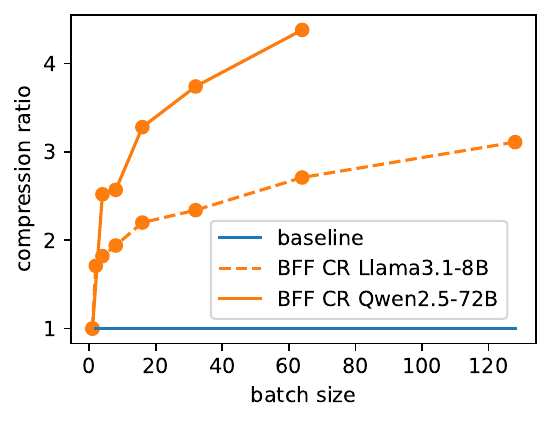}
        \label{fig:bff_math_cr llama qwen}
    }   
    \subfigure[]{
        \includegraphics[width=0.45\textwidth]{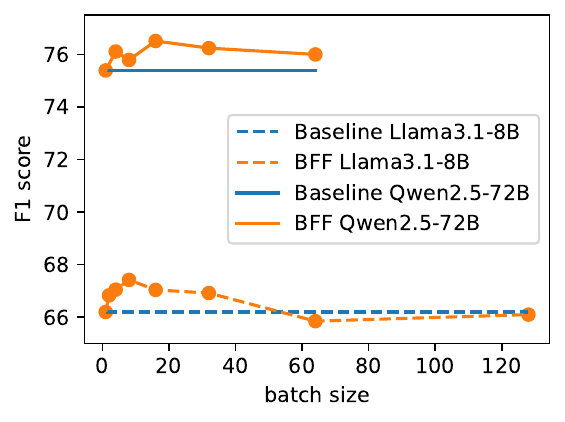}
        \label{fig:bff_math_f1 llama qwen}
   }
   
    \caption{CR and F1 score of BFF vs. baseline for Llama-3.1 8B and Qwen2.5 72B on \citet{nVidiaOpenMathInstruct} dataset. (a) CR vs. batch size.  (b) F1 score vs. batch size.}
    \label{fig:bff_llama3}
\end{figure}

\Cref{fig:bff_math_cr llama qwen} depicts the CR versus batch size for the Llama3.1-8B model and Qwen2.5-72B. Notably, in both cases, the CR grows logarithmically with the batch size, reaching CR of $\sim 3.11\times$ and $\sim 4.38 \times$ for batch sizes of 128 and 64, respectively.

Using the same settings, we further evaluate the behavior of the F1 score when increasing the number of requests in a batch in \Cref{fig:bff_math_f1 llama qwen}. Interestingly, the F1 score is a bit higher on average for both models, and especially for the Qwen2.5 72B model. The phenomenon where averaging similar blocks improves the model accuracy can be interpreted through the lens of the \emph{crowd wisdom effect}. Each block representation, akin to an individual expert, contributes unique insights about the block context over the layers. When focusing on relatively similar representations, averaging these representations reduces individual biases and errors, much as a group makes more accurate decisions \citep{10.1162/opmi_a_00144}.

\Cref{table:bff} describes the F1 score and the CR behavior when applying BFF to every 8 requests (i.e., batch size 8) for various thresholds in a variety of \citet{cais2020mmlu} tasks and \citet{openai2022gsm8k}. These results highlight the ability of the BFF to significantly reduce the KV cache size in many cases, without compromising accuracy, showcasing its value in improving decoding efficiency.
\begin{table*}[ht]
\scriptsize
  \centering 
  \caption{F1 score and CR (in parenthesis) achieved by BFF for Llama3.1-8B.   
  }
  \label{table:bff}
  \begin{tabular}{ccccccccccc}
  \specialrule{.2em}{.1em}{.1em} 
  Model  & Method & GSM8K & Con.Phy. & E.Eng.  & F.Logic &  HS.Bio & misc. & sociology & Average \\
  \specialrule{.2em}{.1em}{.1em} 
     &BFF& 41.47 & 37.87 & \textbf{40.69} & 35.2 & 44.19 & 51.72 & \textbf{40.8} & 41.71 \\
  ~ &thr=0.7& ($\times$3.1) & ($\times$3.29) & ($\times$3.21) & ($\times$2.69) & ($\times$2.51) & ($\times$2.93) & ($\times$3.11) & ($\times$2.98) \\ \cline{2-10}
  & BFF & 51.91 & 40.85 & 38.62 & 40 & 46.13 & 51.34 & 39.3 & 44.02 \\
  Llama3.1-8B &thr=0.74& ($\times$2.52) & ($\times$2.52) & ($\times$2.48) & ($\times$2.21) & ($\times$2.04) & ($\times$2.32) & ($\times$2.4) & ($\times$2.36) \\  \cline{2-10}
  & BFF & \textbf{62.86} & 40.43 & 40 & \textbf{41.27} & 45.48 & \textbf{54.79} & 38.31 & 46.16  \\
  ~ &thr=0.78& ($\times$2.09) & ($\times$1.63) & ($\times$1.59) & ($\times$1.53) & ($\times$1.37) & ($\times$1.52) & ($\times$1.61) & ($\times$1.62) \\  \cline{2-10}
  & Baseline & 62.46 &\textbf{40.85} & 40.69 & 40 & \textbf{48.39} & 54.79 & 40.8 & \textbf{46.85} \\
  \specialrule{.2em}{.1em}{.1em} `
  \end{tabular}
  \normalsize
\end{table*}

Practically, these results indicate up to $\sim 3.11\times$ and $\sim 4.38 \times$  reduction of KV cache blocks, for Llama3.1-8B and Qwen2.5-72B, respectively. Of course, this also mitigates the memory fetching and the network bottleneck which is significant, especially when considering a high level of parallelism. Furthermore, using fused blocks is beneficial for performing hardware-optimized matrix-matrix multiplications \citep{juravsky2024hydragen}.

\vspace{-\baselineskip}
%
   
%
\subsection{Chunks Fast Fusion During Prefill}\label{sec. cff prefill}
In this section, we examine the CR and the resulting F1 score for the CFF. Even though eliminating the distance that stems from RoPE can yield a higher CR, it disables reusing computations, which is substantial for the prefill phase. Thus, the CFF is applied to blocks within chunks, together with their positioning. In addition to compression, the results also indicate the computation reuse factor.

\begin{figure}[t]
    \centering
    \subfigure[]{
        \includegraphics[width=0.45\textwidth]{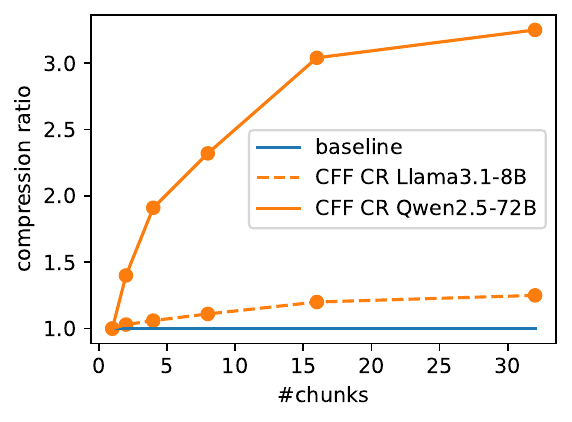}
        \label{fig:cff_cr longbench qmsum}
   }
    \subfigure[]{
        \includegraphics[width=0.45\textwidth]{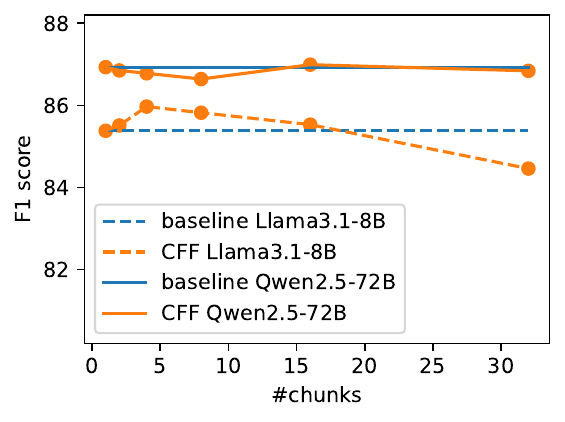}
        \label{fig:cff_f1 longbench qmsum}
    }
   
    \caption{CR (a) and F1 score (b) of CFF vs. number of chunks for Llama3.1-8B and Qwen2.5-72B using the Longbench qmsum dataset. }
    \label{fig:cff_llama3}
\end{figure}

To characterize the impart of CFF on the CR and F1 score, the similarity threshold is set to a fixed value, and the number of chunks to fuse is varied. The evaluation is performed on Llama3.1-8B and Qwen2.5-72B models with 16 tokens per block on the Longbench qmsum dataset \citet{bai2023longbench}. This dataset contains relatively long inputs, which allows characterizing the CR and accuracy of CFF when applied to increasing number of chunks.  \Cref{fig:cff_cr longbench qmsum} depicts the CR and when applying CFF to the chunks.  As we see, the CFF uses up to $\sim3.25\times$ fewer blocks, for which their computation can be reused, thus mitigating computational and network bottlenecks. \Cref{fig:cff_f1 longbench qmsum} depicts the F1 score of the CFF when scaling the number of chunks to fuse on the same task. The CFF manages to keep the accuracy in most cases, and experiences only a negligible accuracy loss. 

\Cref{table:cff} describes the F1 score and the CR behavior when applying CFF to every 8 chunks for various thresholds in a variety of Longbench tasks ~\citep{bai2023longbench}. The table highlights consistent gains in compression with minimal or no loss in accuracy. Notably, even at lower thresholds, the F1 scores remain comparable to the baseline, while achieving compression ratios of up to $\times$1.87.
\begin{table*}[ht]
\scriptsize
  \centering 
  \caption{F1 score and CR (in parenthesis) achieved by CFF for Llama3.1-8B.   
  }
  \label{table:cff}
  \begin{tabular}{ccccccccccc}
  \specialrule{.2em}{.1em}{.1em} 
  Model  & Method & LCC & RepoBench-P & PR-en  & TREC &  2wikimqa & GovReport & MQA-zh & Average \\
  \specialrule{.2em}{.1em}{.1em} 
     &CFF& 75.77 & 74.07 & \textbf{21.04} & 40.48 & \textbf{45.9} & 83.8 & 27.34 & 52.63 \\
  ~ &thr=0.62& ($\times$1.87) & ($\times$1.59) & ($\times$1.32) & ($\times$1.43) & ($\times$1.38) & ($\times$1.59) & ($\times$1.53) & ($\times$1.53) \\ \cline{2-10}
  & CFF & 77.12 & 74.8 & 18.3 & 40.3 & 45.88 & 81.39 & 38.08 & 53.7 \\
  Llama3.1-8B &thr=0.64& ($\times$1.49) & ($\times$1.3) & ($\times$1.15) & ($\times$1.16) & ($\times$1.17) & ($\times$1.25) & ($\times$1.24) & ($\times$1.25) \\  \cline{2-10}
  & CFF & 77.24 & \textbf{75.99} & 18.38 & \textbf{41.15} & 44.94 & \textbf{83.83} & \textbf{40.31} & \textbf{54.55}  \\
  ~ &thr=0.66& ($\times$1.26) & ($\times$1.15) & ($\times$1.06) & ($\times$1.06) & ($\times$1.06) & ($\times$1.09) & ($\times$1.1) & ($\times$1.11) \\  \cline{2-10}
  & Baseline & \textbf{77.73} &75.57 & 18.74 & 40.95 & 44.12 & 82.81 & 39.18 & 54.16 \\
  \specialrule{.2em}{.1em}{.1em} `
  \end{tabular}
  \normalsize
\end{table*}

\subsection{End-to-End Serving Perfromance}\label{sec. e2e performance}
We evaluate BFF on a single-machine vLLM serving benchmark over a random dataset of 100 requests with input size 1024 and varying number of output tokens (with random-ratio 0.5) using Llama-3.1-8B. After the fusing blocks (with similarity above threshold 0.7), the model runner sends the scheduler the updated block-tables, allowing the scheduler to free unused blocks, and increase reference counts for shared blocks for further reuse across requests.

\begin{figure}[t]
    \centering
    \subfigure[]{
        \includegraphics[width=0.45\textwidth]{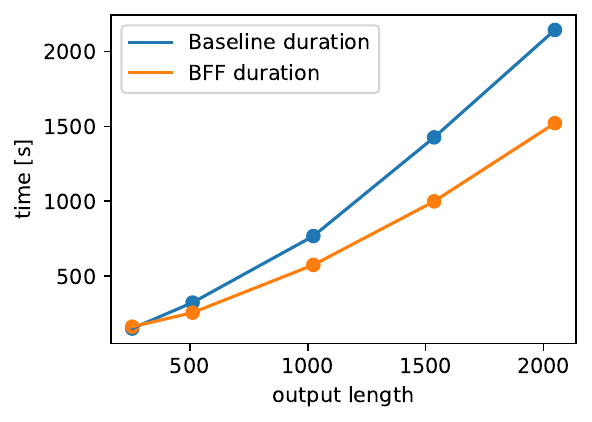}
        \label{fig:benchmark_duration}
   }
    \subfigure[]{
        \includegraphics[width=0.45\textwidth]{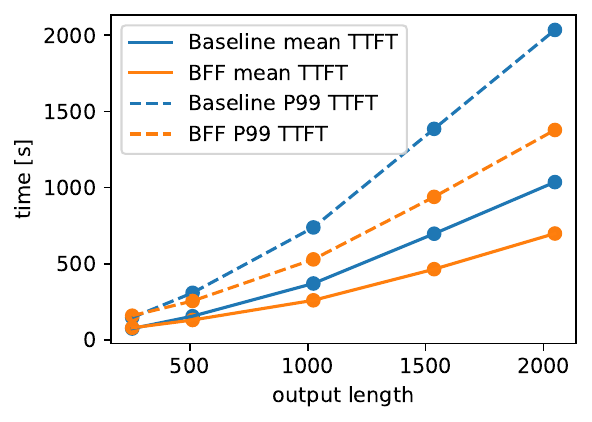}
        \label{fig:BFF_baseline_TTFT}
    }
    \subfigure[]{
        \includegraphics[width=0.45\textwidth]{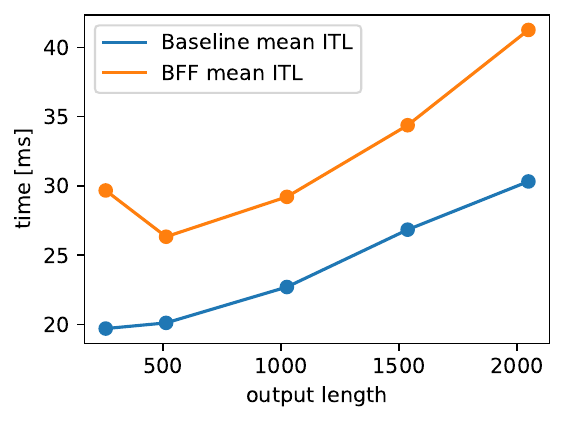}
        \label{fig:BFF_baseline_ITL}
    }
    \subfigure[]{
        \includegraphics[width=0.45\textwidth]{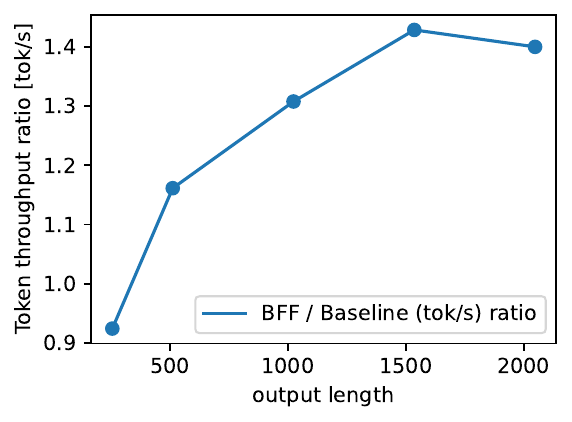}
        \label{fig:BFF_baseline_throuput_ratio}
    }
   
    \caption{Throughput and end-to-end performance comparison on Llama3.1-8B (a) benchmark duration (b) TTFT mean and P99 (c) mean ITL (d) throughput ratio (tok/sec)}
    \label{fig:e2e_performance}
\end{figure}

\Cref{fig:e2e_performance} depicts the throughput and end-o-end performance achieved by our naive implementation, where the entire fusion overhead is reflected in the decode phase. Remarkably, the BFF yields measurable system-level benefits: lower prefill latency (TTFT), higher effective batching, and increased throughput. The Decode-phase inter-token latency (ITL) increased due to larger batch GEMMs, and the fusion overhead. However, the overall serving throughput and request completion time reduced substantially.


Overall, the results presented in this section demonstrate the effectiveness of our context-sharing scheduling scheme for the KV cache-centric disaggregated architecture. Our approach significantly improves compute efficiency, reduces memory and network bandwidth consumption, and scales well with increasing system size, making it a promising solution to accelerate LLM serving in resource-constrained scenarios.

%% file: sections/appendix.tex
\section{Appendix}
\subsection{Proof to \Cref{lem:Attention (softmax) error bound}}
\label{proof:lem attn error}
\begin{proof}
    First, since $\mathbf{k}$ and $\mathbf{k}^\prime$ have the same norm, we can represent then as $\mathbf{k} = \Vert \mathbf{k}\Vert \cdot \mathbf{\hat{k}}$ and $\mathbf{k}^\prime = \Vert \mathbf{k}\Vert \cdot \mathbf{\hat{k}}^\prime$. Let  $\delta_t = \mathbf{z}^\prime_{t} - \mathbf{z}_{t} = \Vert\mathbf{k}\Vert \cdot \mathbf{q}\cdot (\mathbf{\hat{k}}^\prime - \mathbf{\hat{k}})/ \sqrt{d}$. Thus, 
    \begin{align}
        \Vert\delta_t \Vert =& \Vert \mathbf{z}^\prime_{t} - \mathbf{z}_{t}\Vert \\  
        \leq& \Vert \mathbf{k}\Vert \cdot \Vert \mathbf{q} \Vert \cdot \Vert\mathbf{\hat{k}}^\prime - \mathbf{\hat{k}}\Vert / \sqrt{d} \\
        =& \Vert \mathbf{k}\Vert \cdot \Vert \mathbf{q} \Vert \cdot \sqrt{2(1-\mathbf{\hat{k}}^\prime\cdot\mathbf{\hat{k}})} / \sqrt{d} \\
        \leq& \Vert \mathbf{k}\Vert \cdot \Vert \mathbf{q} \Vert \cdot \sqrt{2(1-u)}/ \sqrt{d}      
    \end{align}
    
    Since $\epsilon = \max_i \vert {\delta}_i \vert$, the denominator of $\mathbf{s}$ satisfies
    \[
    \exp(-\epsilon) \sum_j \exp(z_j) \leq \sum_j \exp(z_j + \delta_j) \leq \exp(\epsilon) \sum_j \exp(z_j)
    \]
    Similarly, for the numerator of $\mathbf{s}$ satisfies
    \[
    \exp(-\epsilon) \exp(\mathbf{z}) \leq \exp(\mathbf{z} + \delta) \leq \exp(\epsilon) \exp(\mathbf{z})
    \]
    From these bounds we obtain that each index $i$ satisfies $\exp(-2\epsilon){s}_i \leq {s}_i^\prime \leq \exp(2\epsilon)s$, as 
    \[
    \mathbf{s}^\prime \leq \exp(\epsilon) \exp(\mathbf{z})/\exp(-\epsilon) \sum_j \exp(z_j) = \exp(2\epsilon) \mathbf{s}.
    \]
    Now, let us divide into the case. When $s_i^\prime \geq s_i$, then $s^\prime_i -s_i \leq s_i(\exp(2\epsilon) - 1)$. Similarly, when $s_i^\prime < s_i$, then $s_i - s^\prime_i\leq s_i(1 - \exp(-2\epsilon)) \leq s_i(\exp(2\epsilon) - 1)$. Thus, $\vert s^\prime_i -s_i \vert \leq (\exp(2\epsilon) - 1) s_i$. 

    Finally, sum up the index $i$, since $\sum_i s_i = 1$, we have $\sum_i\vert s^\prime_i -s_i \vert = \Vert \mathbf{s}^\prime - \mathbf{s} \Vert_1 \leq (\exp(2\epsilon) - 1) \leq 2\epsilon + O(\epsilon^2) $ from Taylor expansion.
    Since $\epsilon \leq \max_i \Vert \mathbf{k}_i \Vert \Vert \mathbf{q}\Vert \sqrt{2(1-u)}/\sqrt{d}$, the lemma follows.    
\end{proof}